\documentclass[11pt,a4paper]{article}
\usepackage[hyperref]{acl2017}
\usepackage{times}
\usepackage{latexsym}

\usepackage{url}

\usepackage{amsmath}
\usepackage{multirow}
\usepackage{graphicx}
\usepackage{subcaption}
\usepackage{float}
\usepackage[T1]{fontenc}
\usepackage{color}
\usepackage{comment}

\aclfinalcopy 

\title{Predicting Native Language from Gaze}

\author{Yevgeni Berzak \\
  MIT CSAIL\\
  {\tt berzak@mit.edu} \\\And
  Chie Nakamura \\
  MIT Linguistics \\
  {\tt chienak@mit.edu} \\\And
  Suzanne Flynn \\
  MIT Linguistics \\
  {\tt sflynn@mit.edu} \\\And
  Boris Katz \\
  MIT CSAIL \\
  {\tt boris@mit.edu}
}

\date{}

\begin{document}
\maketitle
\begin{abstract}
A fundamental question in language learning concerns the role of a speaker's first
language in second language acquisition. We present a novel methodology for studying this question: 
analysis of eye-movement patterns in second language reading of free-form text. Using this 
methodology, we demonstrate for the first time that the native language of English learners can be predicted from their gaze 
fixations when reading English. We provide analysis of classifier uncertainty and learned features, 
which indicates that differences in English reading are likely to be rooted in linguistic divergences 
across native languages. The presented framework complements production studies and offers new ground 
for advancing research on multilingualism.\footnote{The experimental data collected in
this study will be made publicly available.}

\end{abstract}

\section{Introduction}

The influence of a speaker's native language on learning and performance in a foreign
language, also known as cross-linguistic transfer, has been studied for several decades 
in linguistics and psychology 
\cite{odlin1989,martohardjono1995,jarvis2008,berkes2012multilingualism,alonso2015}. 
The growing availably of learner corpora has also sparked interest in cross-linguistic influence phenomena
in NLP, where studies have explored the task of Native Language
Identification (NLI) \cite{nli2013report}, as well as analysis of textual features in relation to the 
author's native language \cite{jarvis2012approaching,swanson2013,malmasi2014}.
Despite these advances, the extent and nature of first language influence in second 
language processing remains far from being established. Crucially, most prior work on this topic 
focused on production, while little is currently known about cross-linguistic 
influence in language comprehension. 

In this work, we present a novel framework for studying cross-linguistic influence 
in language comprehension using \emph{eyetracking for reading} and \emph{free-form native English text}. 
We collect and analyze English newswire reading data from 182 participants, including 
145 English as Second Language (ESL) learners from four different native language backgrounds: 
Chinese, Japanese, Portuguese and Spanish, as well as 37 native English speakers. 
Each participant reads 156 English sentences, half of which are shared across all 
participants, and the remaining half are individual to each participant. All the 
sentences are manually annotated with part-of-speech (POS) tags and syntactic 
dependency trees.

We then introduce the task of \emph{Native Language Identification from Reading (NLIR)},
which requires predicting a subject's native language from gaze while reading text in
a second language. Focusing on ESL participants and using a log-linear classifier with word
fixation times normalized for reading speed as features, we obtain 71.03 NLIR accuracy 
in the shared sentences regime. 
We further demonstrate that NLIR can be generalized effectively to the individual  
sentences regime, in which each subject reads a different set of sentences, by grouping fixations according 
to linguistically motivated clustering criteria. In this regime, we obtain an NLIR accuracy of 51.03.

Further on, we provide classification and feature analyses, suggesting 
that the signal underlying NLIR is likely to be related to \emph{linguistic} characteristics
of the respective native languages. First, drawing on previous work on ESL production, 
we observe that classifier uncertainty in NLIR correlates with global linguistic similarities 
across native languages. In other words, the more similar are the languages,
the more similar are the reading patterns of their native speakers in English. 
Second, we perform feature analysis across native and non-native English speakers, and discuss  
structural and lexical factors that could potentially drive some of the non-native reading patterns 
in each of our native languages. 
Taken together, our results provide evidence for a systematic influence of native language properties on reading, and
by extension, on online processing and comprehension in a second language. 

To summarize, we introduce a novel framework for studying cross-linguistic influence in language
learning by using eyetracking for reading free-form English text. We demonstrate
the utility of this framework in the following ways. First, we obtain the first NLIR 
results, addressing both the shared and the individual textual input scenarios. 
We further show that reading preserves linguistic similarities across native languages of 
ESL readers, and perform feature analysis, highlighting key distinctive reading patterns in 
each native language. The proposed framework complements and extends production studies, and 
can inform linguistic inquiry on cross-linguistic influence.

This paper is structured as follows. In section \ref{sec:setup} we present
the data and our experimental setup. Section \ref{sec:nliresl} describes our approach to NLIR
and summarizes the classification results. 
We analyze cross-linguistic influence in reading in section \ref{sec:transfer}. 
In section \ref{subsec:comparingesl} we examine NLIR classification uncertainty in relation 
to linguistic similarities between native languages. In section \ref{sec:features} 
we discuss several key fixation features associated with different native languages.
Section \ref{sec:relatedwork} surveys related work, and section \ref{sec:conclusion} concludes.

\section{Experimental Setup}
\label{sec:setup}

\subsection*{Participants}

We recruited 182 adult participants. Of those, 37 are native English speakers
and 145 are ESL learners from four native language backgrounds: Chinese, Japanese, 
Portuguese and Spanish. All the participants in the experiment are native speakers of only one language. 
The ESL speakers were tested for English proficiency using the grammar and listening 
sections of the Michigan English test (MET), which consist of 50 multiple choice questions. 
The English proficiency score was calculated as the number of correctly answered questions 
on these modules. The majority of the participants scored in the intermediate-advanced 
proficiency range. Table \ref{table-participants} presents the number of participants and the mean
English proficiency score for each native language group. Additionally, we collected 
metadata on gender, age, level of education, duration of English studies and usage, 
time spent in English speaking countries and proficiency in any additional language spoken. 

\begin{table}[h]
\centering
\begin{tabular}{lcc}
         & \bf \# Participants & \bf English Score \\ \hline
Chinese  &  36 & 42.0\\
Japanese &  36 & 40.3 \\
Portuguese & 36 & 41.1 \\ 
Spanish  &  37 & 42.4 \\
English  &  37  & NA \\ \hline 

\end{tabular}
\caption{Number of participants and mean MET English score by native language group.}
\label{table-participants}
\end{table}

\subsection*{Reading Materials}

We utilize 14,274 randomly selected sentences from the Wall 
Street Journal part of the Penn Treebank (WSJ-PTB) \cite{marcus1993}.
To support reading convenience and measurement precision, the maximal 
sentence length was set to 100 characters, leading to an average sentence 
length of 11.4 words. Word boundaries are defined as whitespaces.
From this sentence pool, 78 sentences (900 words) were presented to all 
participants (henceforth \emph{shared} sentences) and the remaining 14,196 sentences 
were split into 182 individual batches of 78 sentences
(henceforth \emph{individual} sentences, averaging 880 words per batch). 

All the sentences include syntactic annotations from the Universal Dependency 
Treebank project (UDT) \cite{mcdonald2013universal}. The annotations include PTB POS tags 
\cite{ptbpos}, Google universal POS tags \cite{petrov2012} and dependency 
trees. The dependency annotations of the UDT are converted automatically from 
the manual phrase structure tree annotations of the WSJ-PTB.

\subsection*{Gaze Data Collection}
Each participant read 157 sentences. The first sentence was presented to familiarize 
participants with the experimental setup and was discarded during analysis. 
The following 156 sentences consisted of 78 shared and 78 individual sentences. The shared and 
the individual sentences were mixed randomly and presented to all participants in the same order.
The experiment was divided into three parts, consisting of 52 sentences each. Participants were 
allowed to take a short break between experimental parts. 

Each sentence was presented on a blank screen as a one-liner. 
The text appeared in Times font, with font size 23. 
To encourage attentive reading, upon completion of sentence reading participants answered 
a simple yes/no question about its content, and were subsequently informed if they answered the question correctly. 
Both the sentences and the questions were triggered by a 300ms gaze on a fixation target 
(fixation circle for sentences and the letter ``Q'' for questions) which appeared on a 
blank screen and was co-located with the beginning of the text in the following screen.

Throughout the experiment, participants held a joystick with buttons for indicating
completion of sentence reading and answering the comprehension questions.
Eye-movement of participants' dominant eye was recorded using a desktop mount Eyelink 1000 eyetracker, at a sampling rate of 1000Hz.
Further details on the experimental setup are provided in appendix \ref{sec:supplemental}. 

\section{Native Language Identification from Reading}
\label{sec:nliresl}
Our first goal is to determine whether the native language of ESL
learners can be decoded from their gaze patterns while reading English text. 
We address this question in two regimes, corresponding to 
our division of reading input into shared and individual sentences.
In the \emph{shared regime}, all the participants read the same set of  
sentences. Normalizing over the reading input, this regime facilitates focusing
on differences in reading behavior across readers. In the \emph{individual regime}, 
we use the individual batches from our data to address the more challenging variant 
of the NLIR task in which the reading material given to each participant is different.

\subsection{Features}
We seek to utilize features that can provide robust, simple and 
interpretable characterizations of reading patterns. To this end, we use 
speed normalized \emph{fixation duration} measures over word sequences. 

\subsubsection*{Fixation Measures}
We utilize three measures of word fixation duration:
\begin{itemize}
\item \emph{First Fixation duration (FF)} Duration of the first fixation on a word.
\item \emph{First Pass duration (FP)} Time spent from first entering a word to first 
leaving it (including re-fixations within the word). 
\item \emph{Total Fixation duration (TF)} The sum of all fixation times on a word.
\end{itemize}

We experiment with fixations over unigram, bigram and trigram sequences
$seq_{i,k} = w_{i},...,w_{i+k-1}, k \in \{1,2,3\}$, where for each metric $M \in \{FF, FP, TF\}$ the fixation 
time for a sequence $M_{seq_{i,k}}$ is defined as the sum of fixations on individual tokens $M_{w}$ in the sequence\footnote{Note that for bigrams and trigrams, one could
also measure FF and FP for interest regions spanning the sequence, instead, or in addition to summing these fixation
times over individual tokens.}.
\begin{equation}
M_{seq_{i,k}} = \sum_{w' \in seq_{i,k}}M_{w'}
\end{equation}

Importantly, we control for variation in reading speeds across subjects by normalizing 
each subjects's sequence fixation times. For each metric $M$ and sequence $seq_{i,k}$ 
we normalize the sequence fixation time $M_{seq_{i,k}}$ relative to the subject's sequence fixation times in %
the textual context of the sequence. The context $C$ is defined as the sentence in which the sequence appears for 
the \emph{Words in Fixed Context} feature-set and the entire textual input for the \emph{Syntactic} and \emph{Information} clusters 
feature-sets (see definitions of feature-sets below). The normalization term $S_{M,C,k}$ is accordingly defined as the metric's 
fixation time per sequence of length $k$ in the context: 
\begin{equation}
S_{M,C,k} = \frac{1}{|C|}\sum_{seq_{k} \in C}{M_{seq_{k}}}
\end{equation}

We then obtain a normalized fixation time $Mnorm_{seq_{i,k}}$ as:
\begin{equation}
Mnorm_{seq_{i,k}} = \frac{M_{seq_{i,k}}}{S_{M,C,k}}
\end{equation}

\subsubsection*{Feature Types}
We use the above presented speed normalized fixation metrics to extract three feature-sets, 
\emph{Words in Fixed Context (WFC)}, \emph{Syntactic Clusters} (SC) and \emph{Information Clusters (IC)}. 
WFC is a token-level feature-set that presupposes a fixed textual input for all participants. It is thus applicable 
only in the shared sentences regime. SC and IC are type-level features which provide abstractions over sequences of 
words. Crucially, they can also be applied when participants read different sentences.

\begin{itemize}
\item \textbf{Words in Fixed Context (WFC)}
The WFC features capture fixation times on word sequences in a specific sentence. This feature-set consists of
FF, FP and TF times for each of the 900 unigram, 822 bigram, and 744 trigram word sequences comprising the shared sentences. 
The fixation times of each metric are normalized for each participant relative 
to their fixations on sequences of the same length in the surrounding sentence. 
As noted above, the WFC feature-set is not applicable in the individual regime, 
as it requires identical sentences for all participants. 

\item \textbf{Syntactic Clusters (SC)} CS features are average globally normalized FF, FP and TF times for word sequences clustered 
by our three types of syntactic labels: universal POS, PTB POS, and syntactic relation labels. 
An example of such a feature is the average of speed-normalized TF times spent on the PTB POS bigram
sequence DT NN. We take into account labels that appear at least once in the reading input of all participants. 
On the four non-native languages, considering all three label types, we obtain 104 unigram, 
636 bigram and 1,310 trigram SC features per fixation metric in the shared regime, and 56 unigram, 95 bigram and 43 
trigram SC features per fixation metric in the individual regime.

\item \textbf{Information Clusters (IC)} We also obtain average FF, FP and TF for words  
clustered according to their \emph{length}, measured in number of characters. Word length was previously shown to 
be a strong predictor of \emph{information content} 
\cite{piantadosi2011}. As such, it provides an alternative abstraction to the syntactic clusters, combining
both syntactic and lexical information. As with SC features, we take into account features that appear
at least once in the textual input of all participants. For our set of non-native languages, we obtain for each fixation metric
15 unigram, 21 bigram and 23 trigram IC features in the shared regime, and 12 unigram, 18 bigram and 18 trigram IC features in the individual regime. 
Notably, this feature-set is very compact, and differently from the syntactic clusters, does not 
rely on the availability of external annotations.
\end{itemize}

In each feature-set, we perform a final preprocessing step for each individual feature, 
in which we derive a zero mean unit variance scaler from the training set feature values, and apply it to transform both the 
training and the test values of the feature to Z scores.

\subsection{Model}
The experiments are carried out using a log-linear model:
\begin{equation}
p(y|x;\theta) = \frac{\exp(\theta \cdot f(x,y))}{\sum_{y' \in Y} \exp(\theta \cdot f(x,y'))}
\end{equation}
where $y$ is the reader's native language, $x$ is the reading input and 
$\theta$ are the model parameters. 
The classifier is trained with gradient descent using L-BFGS \cite{lbfgs1995}.    
\subsection{Experimental Results}

\begin{table*}[ht]
\resizebox{\textwidth}{!}{
\begin{tabular}{l|ccc|ccc}
                 &  \multicolumn{3}{c|}{\bf Shared Sentences Regime} & \multicolumn{3}{c}{\bf Individual Sentences Regime} \\ \hline
\bf Majority Class & \multicolumn{3}{c|}{25.52} & \multicolumn{3}{c}{25.52} \\ 
\bf Random Clusters & \multicolumn{3}{c|}{22.76} & \multicolumn{3}{c}{22.07}\\ \hline
 & unigrams & +bigrams & +trigrams & unigrams & +bigrams & +trigrams \\ \hline
\bf Information Clusters (IC) 	& 41.38 & 44.14 & 46.21 & 38.62 & 32.41 & 32.41\\
\bf Syntactic Clusters (SC) 	& 45.52 & 57.24 & 58.62	& 48.97 & 43.45 & 48.28 \\
\bf SC+IC 			& 51.72	& 57.24 & 60.0 & \bf 51.03 & 46.21 &  49.66 \\ \hline
\bf Words in Fixed Context (WFC)& 64.14 & 68.28 & \bf 71.03 & \multicolumn{3}{c}{NA} \\\hline
\end{tabular}
}
\caption{\label{nli-table} Native Language Identification from Reading results with 10-fold cross-validation
for native speakers of Chinese, Japanese, Portuguese and Spanish. In the \emph{Shared}
regime all the participants read the same 78 sentences. In the \emph{Individual} regime each
participant reads a different set of 78 sentences.}
\end{table*}

In table \ref{nli-table} we report 10-fold cross-validation results on NLIR in the 
shared and the individual experimental regimes for native speakers of Chinese, Japanese, 
Portuguese and Spanish. We introduce two baselines against which we compare the 
performance of our feature-sets. The \emph{majority} baseline selects the native 
language with the largest number of participants. The \emph{random clusters} baseline 
clusters words into groups randomly, with the number of groups set to the number
of syntactic categories in our data. 

In the shared regime, WFC fixations yield the highest 
classification rates, substantially outperforming the cluster feature-sets
and the two baselines. The strongest result using this feature-set, 71.03, is obtained 
by combining unigram, bigram and trigram fixation times. In addition to this outcome, we 
note that training binary classifiers in this setup yields accuracies ranging from 68.49 
for the language pair Portuguese and Spanish, to 93.15 for Spanish and Japanese.
These results confirm the effectiveness of the shared input regime for performing reliable NLIR, 
and suggest a strong native language signal in non-native reading fixation times.

SC features yield accuracies of 45.52 to 58.62 on the shared sentences, while IC 
features exhibit weaker performance in this regime, with accuracies of 41.38 to 46.21. 
Both results are well above chance, but lower than WFC fixations due to the information loss 
imposed by the clustering step. Crucially, both feature-sets remain effective in the individual 
input regime, with 43.45 to 48.97 accuracy for SC features and 32.41 to 38.62 accuracy for IC features. 
The strongest result in the individual regime is 51.03, obtained by concatenating IC and SC features
over unigrams. We also note that using this setup in a binary classification scheme yields 
results ranging from chance level 49.31 for Portuguese versus Spanish, to 84.93 on Spanish 
versus Japanese.

Generally, we observe that adding bigram and trigram fixations in the shared regime leads to 
performance improvements compared to using unigram features only. This trend
does not hold for the individual sentences, presumably due to a combination 
of feature sparsity and context variation in this regime. We also note 
that IC and SC features tend to perform better together than in separation, 
suggesting that the information encoded 
using these feature-sets is to some extent complementary.

The generalization power of our cluster based feature-sets has both practical 
and theoretical consequences. Practically, they provide 
useful abstractions for performing NLIR over arbitrary textual input. That is,
they enable performing this task using \emph{any} textual input during both training
and testing phases. Theoretically, the effectiveness of linguistically motivated 
features in discerning native languages suggests that linguistic factors play an important role 
in the ESL reading process. The analysis presented in the following sections will further explore 
this hypothesis. 

\section{Analysis of Cross-Linguistic Influence in ESL Reading}
\label{sec:transfer}

As mentioned in the previous section, the ability to perform NLIR in general, and the effectiveness of 
linguistically motivated features in particular, suggest that linguistic factors in the native 
and second languages are pertinent to ESL reading. In this section we explore this hypothesis 
further, by analyzing classifier uncertainty and the features learned in the NLIR task. 

\subsection{Preservation of Linguistic Similarity} 
\label{subsec:comparingesl}

Previous work in NLP suggested a link between textual patterns in ESL production and linguistic similarities 
of the respective native languages \cite{nagata13,nagata14,berzak2014,berzak2015}. In particular, 
\newcite{berzak2014} has demonstrated that NLI classification uncertainty correlates with 
similarities between languages with respect to their typological features. 
Here, we extend this framework and examine if preservation of native language similarities in 
ESL production is paralleled in reading. 

Similarly to \newcite{berzak2014} we define the classification uncertainty for a pair 
of native languages $y$ and $y'$ in our data collection $D$, as the average probability assigned by the NLIR 
classifier to one language given the other being the true native language. This approach provides a robust 
measure of classification confusion that does not rely on the actual performance of the classifier.
We interpret the classifier uncertainty as a similarity measure between the respective 
languages and denote it as English Reading Similarity $ERS$. 

\begin{equation}
\resizebox{\columnwidth}{!}{$ERS_{y,y'} = \frac{\sum\limits_{(x, y) \in D_{y}}p(y'|x;\theta)+\sum\limits_{(x, y') \in D_{y'}} p(y|x;\theta)}{\left\vert{D_{y}}\right\vert + \left\vert{D_{y'}}\right\vert}$}
\label{simeq}
\end{equation}

We compare these reading similarities to the linguistic similarities between
our native languages. To approximate these similarities, we utilize feature vectors from the
URIEL Typological Compendium \cite{uriel} extracted using the \emph{lang2vec} tool \cite{littell2017lang2vec}. 
URIEL aggregates, fuses and normalizes typological, phylogenetic and geographical information about the world's languages. 

We obtain all the 103 available morpho-syntactic features in URIEL, which
are derived from the World Atlas of Language Structures (WALS) \cite{wals}, Syntactic Structures of the World's Languages (SSWL) 
\cite{sswl} and Ethnologue \cite{ethnologue-2015}. Missing 
feature values are completed with a KNN classifier. We also extract URIEL's 3,718 
language family features derived from Glottolog \cite{hammarstrom2015glottolog}. Each of these features 
represents membership in a branch of Glottolog's world language tree. Truncating features with 
the same value for all our languages, we remain with 76 features, consisting of 49 syntactic features 
and 27 family tree features. The linguistic similarity $LS$ between a pair of languages $y$ and $y'$ is then
determined by the cosine similarity of their URIEL feature vectors. 
\begin{equation}
LS_{y,y'} = \frac{v_{y} \cdot v_{y'}}{\Vert v_{y}\Vert \Vert v_{y'}\Vert }
\end{equation}

\begin{figure}[H]
    \centering
    \begin{subfigure}[b]{\columnwidth}
        \includegraphics[width=1.0\columnwidth]{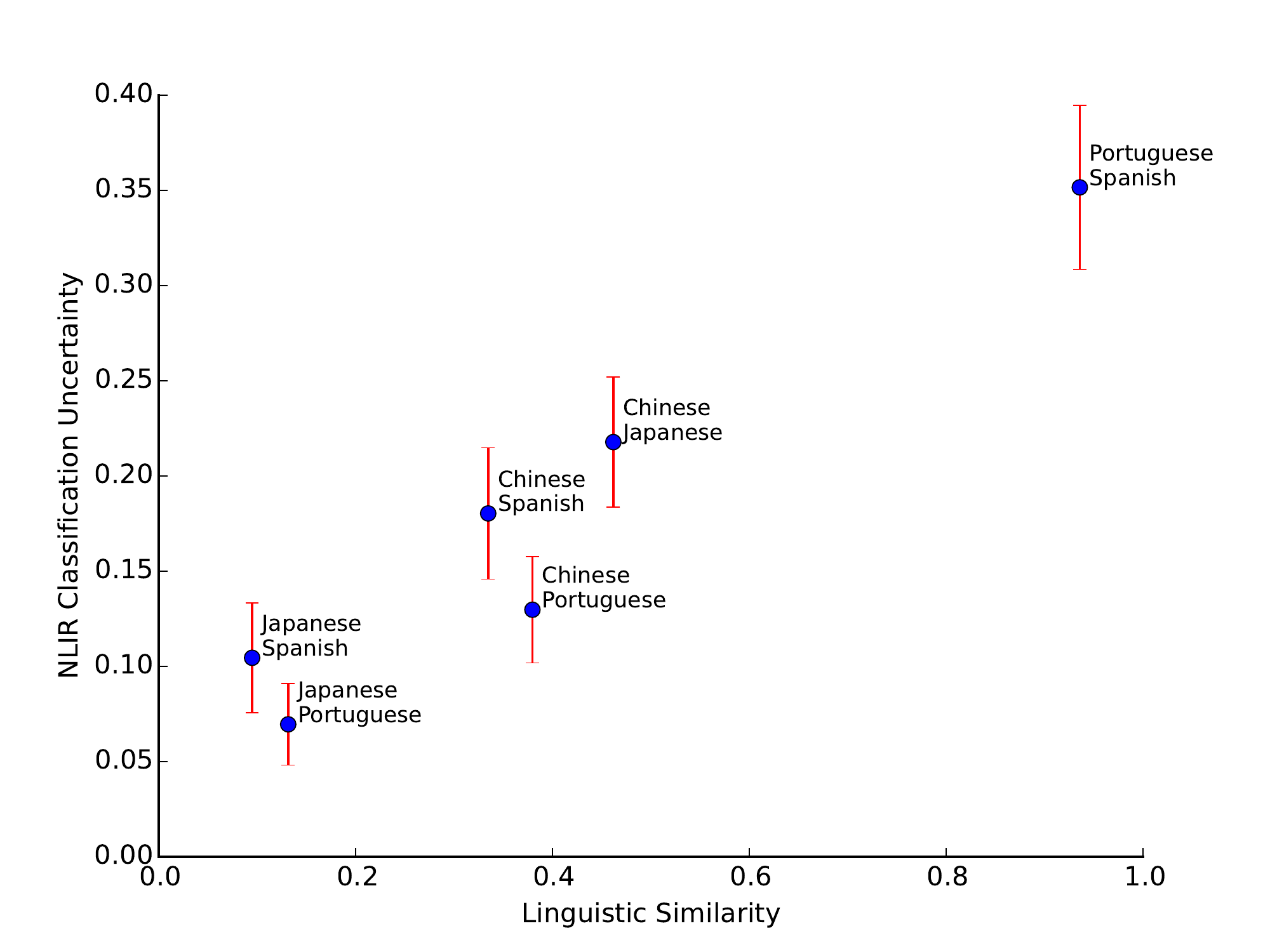}
        \caption{Linguistic similarities against mean NLIR classification uncertainty. Error bars denote standard error.}
        \label{fig:uriel-vs-classifier}
    \end{subfigure} \\
\begin{subfigure}[b]{0.47\columnwidth}
        \includegraphics[width=\columnwidth]{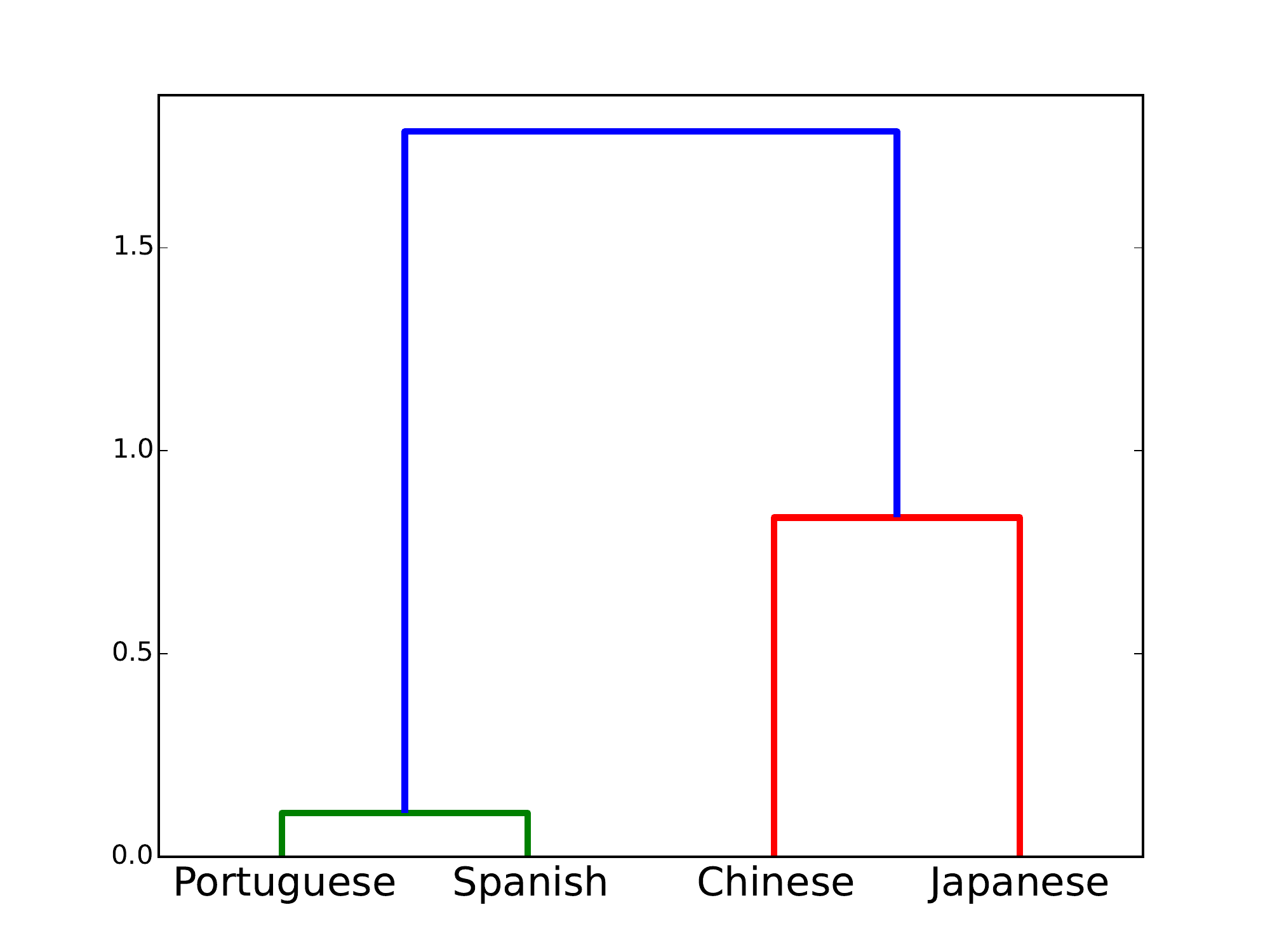}
        \caption{Linguistic tree}
        \label{fig:uriel-tree}
    \end{subfigure}
    \begin{subfigure}[b]{0.47\columnwidth}
        \includegraphics[width=\columnwidth]{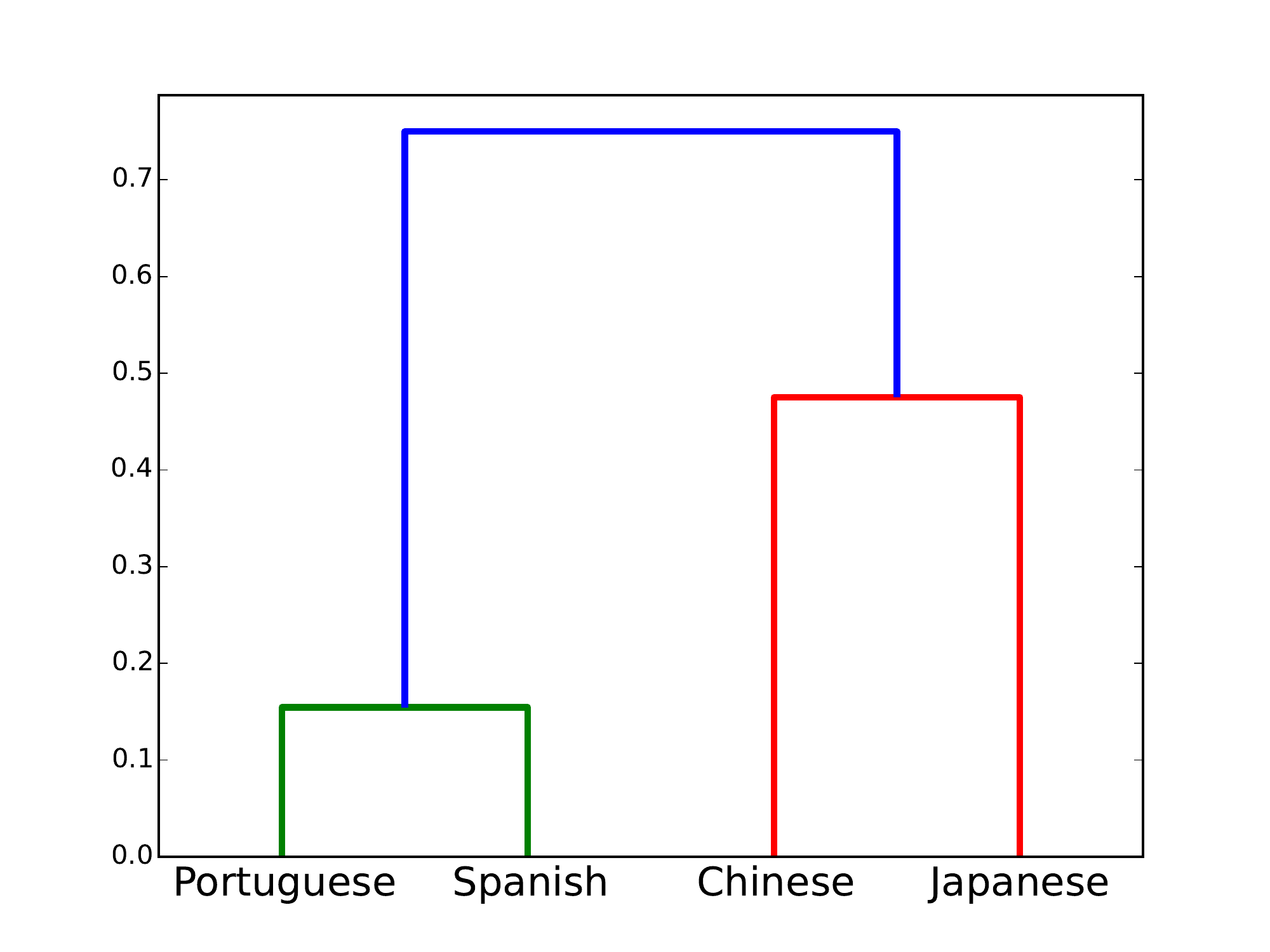}
        \caption{English reading tree}
        \label{fig:classifier-tree}
    \end{subfigure}
    \caption{ (a) Linguistic versus English reading language similarities. 
                 The horizontal axis represents typological and phylogenetic similarity between languages, 
 		 obtained by vectorizing linguistic features form URIEL, and measuring their 
		 cosine similarity. The vertical axis represents the average uncertainty of the NLIR 
		 classifier in distinguishing ESL readers of each language pair. 
		 (b) Ward hierarchical clustering of linguistic similarities between languages.
		 (c) Ward hierarchical clustering of NLIR average pairwise classification uncertainties.
	     }
\label{fig:esltransfer}
\end{figure}

Figure \ref{fig:esltransfer} presents the URIEL based linguistic similarities for our 
set of non-native languages against the average NLIR classification uncertainties on the 
cross-validation test samples. The results presented in this figure are based on the unigram 
IC+SC feature-set in the individual sentences regime. We also provide a graphical illustration 
of the language similarities for each measure, using the Ward clustering algorithm \cite{ward1963}. 
We observe a correlation between the two measures which is also reflected in similar hierarchies
in the two language trees. Thus, linguistically motived features in English reveal 
linguistic similarities across native languages. This outcome supports the hypothesis that 
English reading differences across native languages are related to linguistic factors.

We note that while comparable results are obtained for the IC and SC feature-sets, together and in separation 
in the shared regime, WFC features in the shared regime do not exhibit a clear uncertainty distinction when comparing across the pairs 
Japanese and Spanish, Japanese and Portuguese, Chinese and Spanish, and Chinese and Portuguese. Instead, this 
feature-set yields very low uncertainty, and correspondingly very high performance ranging from 90.41 to 93.15, 
for all four language pairs.

\subsection{Feature Analysis}
\label{sec:features}

Our framework enables not only native language classification,
but also exploratory analysis of native language specific reading patterns 
in English. The basic question that we examine in this respect is on which 
features do readers of different native language groups spend more versus
less time. We also discuss several potential relations of the observed reading time 
differences to usage patterns and grammatical errors committed by speakers of our 
four native languages in production. We obtain this information by extracting 
grammatical error counts from the CLC FCE corpus \cite{fce2011}, and from the ngram 
frequency analysis in Nagata and Whittaker \shortcite{nagata13}.

In order to obtain a common benchmark for reading time comparisons across non-native
speakers, in this analysis we also consider our group of native English speakers.
In this context, we train four binary classifiers that discern each of the non-native groups from 
native English speakers based on TF times over unigram PTB POS tags in the shared regime. 
The features with the strongest positive and negative weights learned by these classifiers
are presented in table \ref{features-table}. These features 
serve as a reference point for selecting the case studies discussed below.

\begin{figure}[H]
    \centering
\begin{subfigure}[b]{0.48\columnwidth}
        \includegraphics[width=\columnwidth]{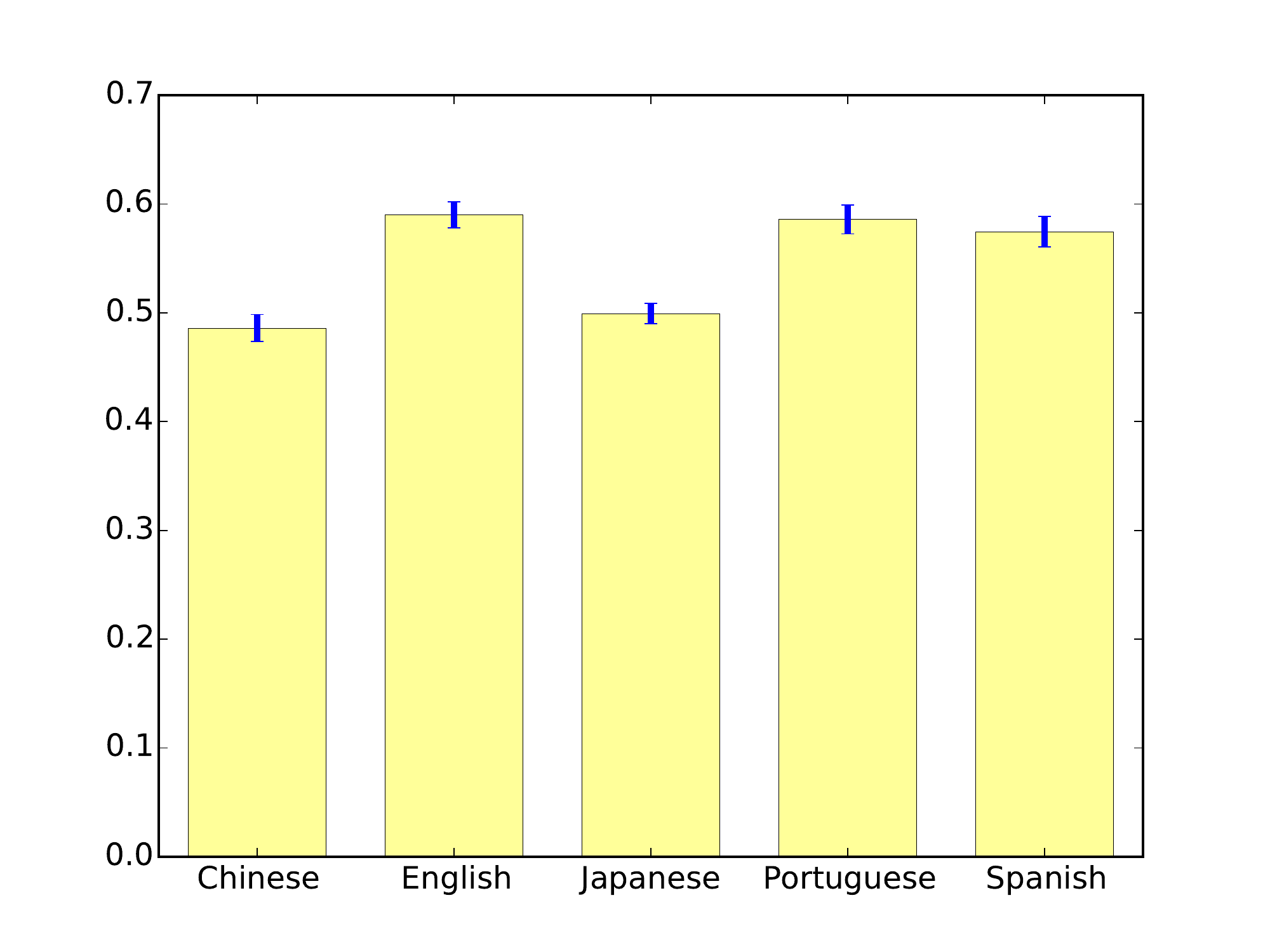}
        \caption{Determiners (DT)}
        \label{fig:determiners}
    \end{subfigure}
    \begin{subfigure}[b]{0.48\columnwidth}
        \includegraphics[width=\columnwidth]{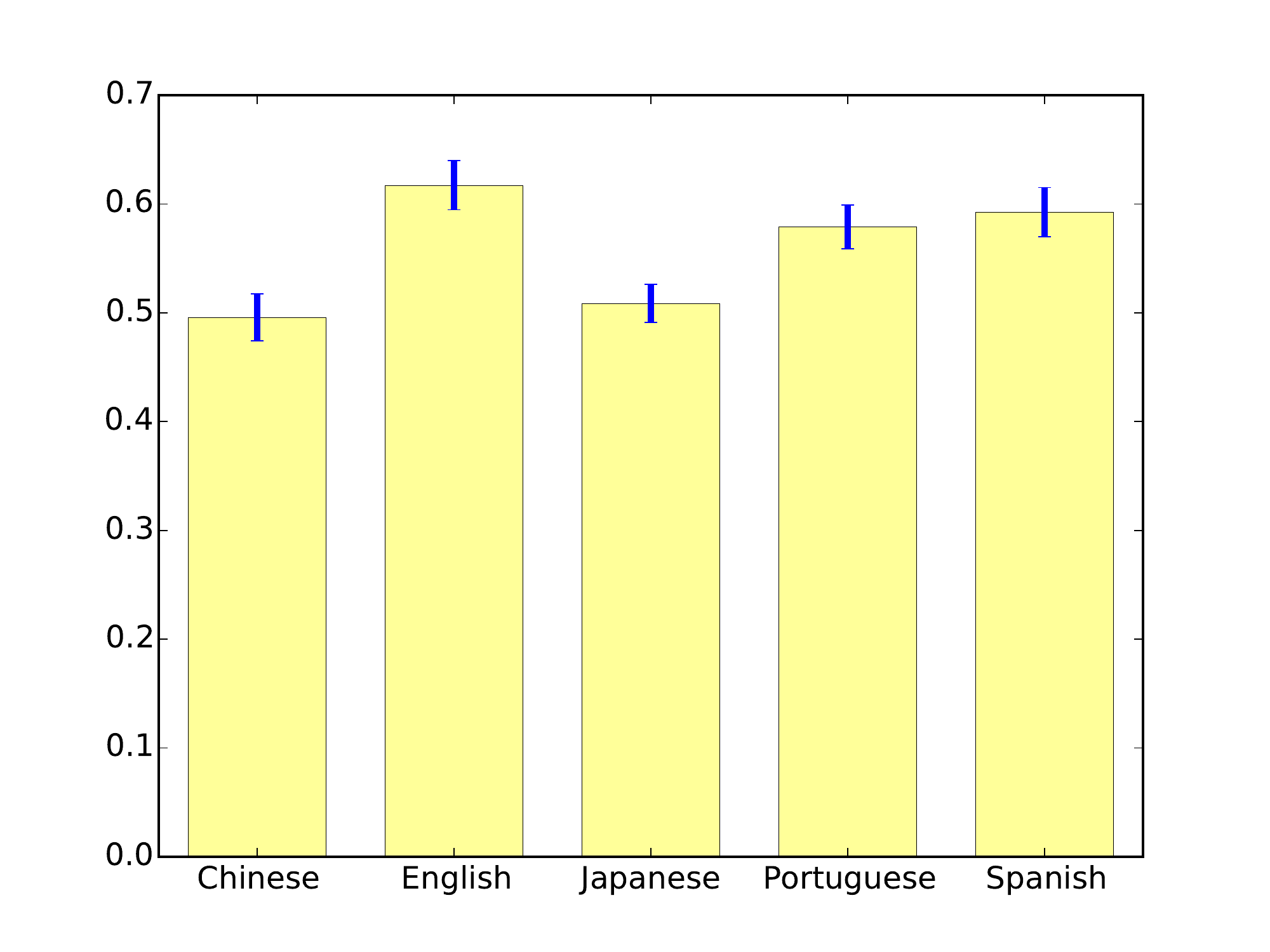}
        \caption{Pronouns (PRP)}
        \label{fig:pronouns}
    \end{subfigure} \\

    \begin{subfigure}[b]{0.48\columnwidth}
        \includegraphics[width=\columnwidth]{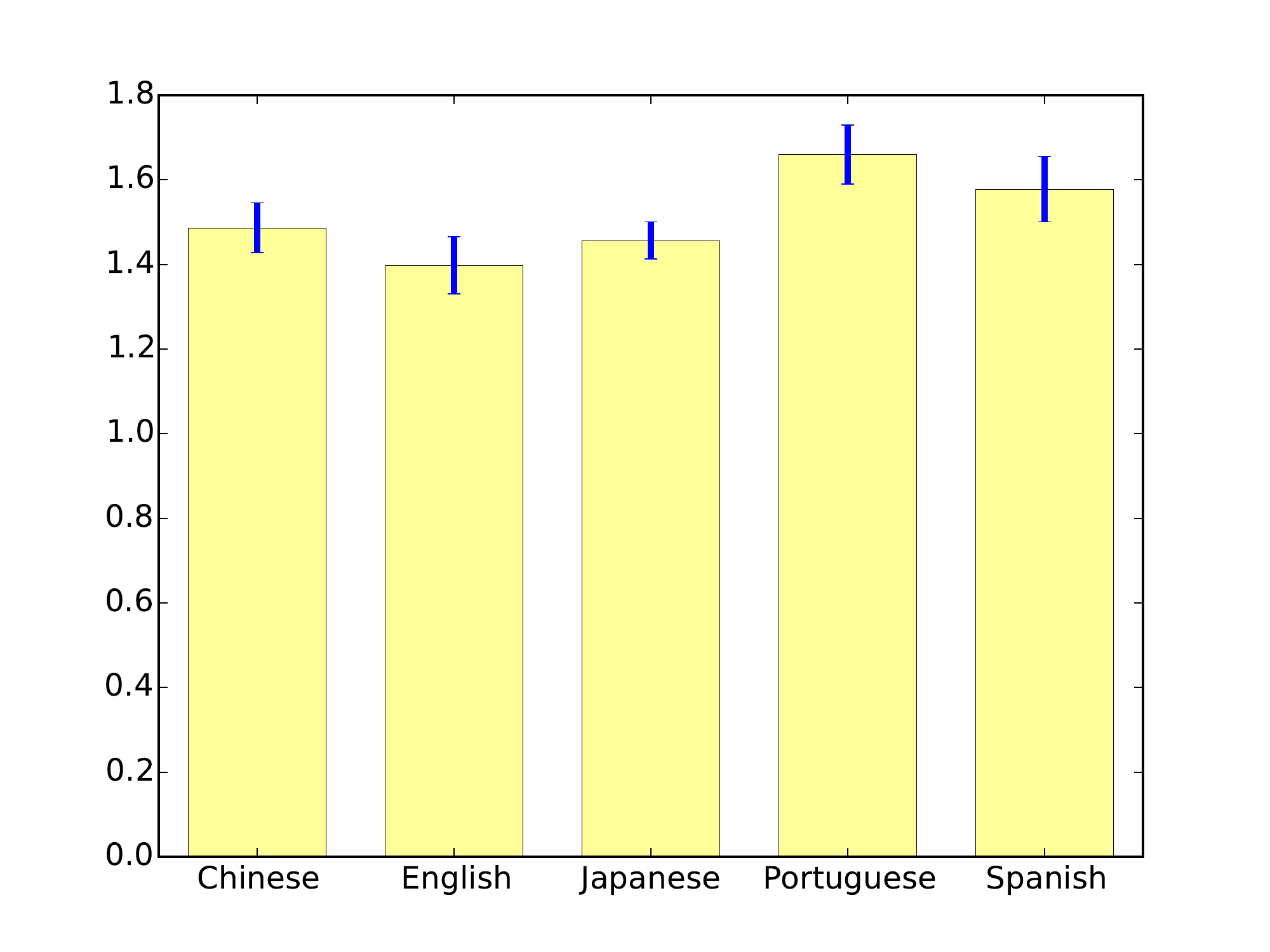}
        \caption{Possessives (NN+POS)}
        \label{fig:possessives}
    \end{subfigure}
    \begin{subfigure}[b]{0.48\columnwidth}
        \includegraphics[width=\columnwidth]{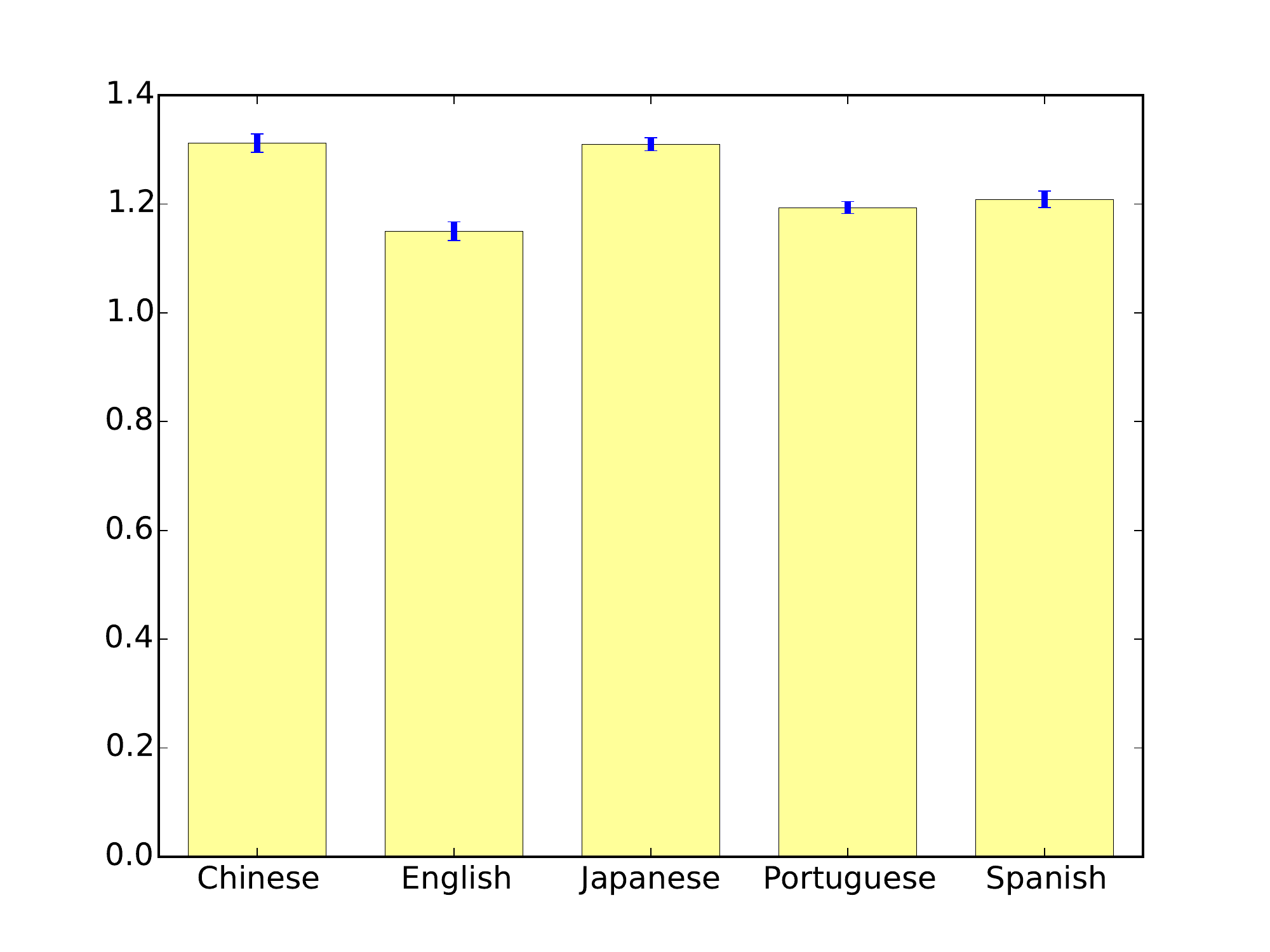}
        \caption{Nouns (NN)}
        \label{fig:nn}
    \end{subfigure}
    \caption{Mean speed-normalized Total Fixation duration for Determiners (DT), Pronouns (PRP),
singular noun possessives (NN+POS), and singular nouns (NN) appearing in the shared sentences. Error bars denote standard error. 
	     }
\label{fig:bars}
\end{figure}

Interestingly, some of the reading features that are most 
predictive of each native language lend themselves to linguistic interpretation with respect
to \emph{structural} factors.
For example, in Japanese and Chinese we observe shorter reading times for determiners (DT), 
which do not exist in these languages. Figure \ref{fig:determiners} presents the mean TF 
times for determiners in all five native languages, suggesting that native speakers of 
Portuguese and Spanish, which do have determiners, do not exhibit reduced reading times 
on this structure compared to natives. In ESL production, missing determiner errors are the most
frequent error for native speakers of Japanese and third most common error for native speakers
of Chinese. 

In figure \ref{fig:pronouns} we present the mean TF reading times for pronouns (PRP),
where we also see shorter reading times by natives of Japanese and Chinese as compared to English
natives. In both languages pronouns can be omitted both in object and subject positions. 
Portuguese and Spanish, in which pronoun omission is restricted to the subject position 
present similar albeit weaker tendency.

\begin{table}[!ht]
\small
\begin{tabular}{lll}
 & \bf Negative (Fast)  & \bf Positive (Slow) \\ \hline
\bf Chinese  	& \bf DT	& JJR \\ 
	     	& \bf PRP 	& \bf NN \\ \hline
\bf Japanese 	& \bf DT   	& \bf NN \\ 
             	&  CD  	& VBD  \\  \hline
\bf Portuguese  & NNS		& \bf NN-POS \\ 
		& \bf PRP	&  VBZ \\ \hline
\bf Spanish     &  NNS 		&  MD \\
	        & \bf PRP    	&  RB \\ \hline
\end{tabular}
\caption{\label{features-table} PTB POS features with the strongest weights learned in non-native versus 
				native classification for each native language in the shared regime. 
				Feature types presented in figure \ref{fig:bars} are highlighted in bold.}
\end{table} 

In figure \ref{fig:possessives} we further observe that differently from natives of Chinese and Japanese, 
native speakers of Portuguese and Spanish spend more time on NN+POS in head final possessives such as 
``the \emph{public's} confidence''. While similar constructions exist in Chinese and Japanese, 
the NN+POS combination is expressed in Portuguese and Spanish as a head initial NN \emph{of NN}. 
This form exists in English (e.g. ``the confidence of the public'') and is preferred by speakers of these languages in ESL writing \cite{nagata13}. 
As an additional baseline for this construction, we provide the TF times for NN in figure \ref{fig:nn}. There,
relative to English natives, we observe longer reading times for Japanese and Chinese and comparable 
times for Portuguese and Spanish.

The reading times of NN in figure \ref{fig:nn} also give rise to a second, potentially competing 
interpretation of differences in ESL reading times, which highlights \emph{lexical} rather than 
structural factors. According to this interpretation, increased reading times of nouns  
are the result of substantially smaller lexical sharing with English by Chinese and Japanese 
as compared to Spanish and Portuguese. Given the utilized speed normalization, 
lexical effects on nouns could in principle account for reduced reading 
times on determiners and pronouns. Conversely, structural influence leading to reduced reading 
times on determiners and pronouns could explain longer dwelling on nouns. A third possibility consistent with the
observed reading patterns would allow for both structural and lexical effects to impact second
language reading. Importantly, in each of these scenarios, ESL 
reading patterns are related to linguistic factors of the reader's native language. 

We note that the presented analysis is preliminary in nature, and warrants further study
in future research. In particular, reading times and classifier learned features may in some 
cases differ between the shared and the individual regimes. In the examples presented above, 
similar results are obtained in the individual sentences regime for DT, PRP and NN. The trend
for the NN+POS construction, however, diminishes in that setup with similar reading 
times for all languages. On the other hand, one of the strongest features for predicting Portuguese and Spanish in the individual regime
are longer reading times for prepositions (IN), an outcome that holds in the shared regime only
relative to Chinese and Japanese, but not relative to native speakers of English.  

Despite these caveats, our results suggest that reading patterns can potentially be related 
to linguistic factors of the reader's native language. This analysis can be extended in various ways, such as inclusion
of additional feature types and fixation metrics, as well as utilization
of other comparative methodologies. Combined with evidence from language production, this line of investigation can 
be instrumental for informing linguistic theory of cross-linguistic influence.

\section{Related Work}
\label{sec:relatedwork}

\textbf{Eyetracking and second language reading} Second language reading has been studied using eyetracking, 
with much of the work focusing on processing of syntactic ambiguities and analysis of specific target word 
classes such as cognates \cite{dussias2010uses,roberts2013using}. In contrast to our work, such studies typically
use controlled, rather than free-form sentences. Investigation of global metrics in 
free-form second language reading was introduced only recently by Cop et al. \shortcite{novel15}. 
This study compared ESL and native reading of a novel by native speakers of Dutch, observing longer 
sentence reading times, more fixations and shorter saccades in ESL reading. Differently from this study, our work focuses 
on comparison of reading patterns between different native languages. We also analyze a related, 
but different metric, namely speed normalized fixation durations on word sequences.

\textbf{Eyetracking for NLP tasks} Recent work in NLP has demonstrated that reading gaze can serve as 
a valuable supervision signal for standard NLP tasks. Prominent examples of such work include POS tagging 
\cite{barrett2015pos,barrett2016pos}, syntactic parsing \cite{barrett2015parsing} and sentence compression 
\cite{klerke2016compression}. Our work also tackles a traditional NLP task with free-form text, but 
differs from this line of research in that it addresses this task only in comprehension. 
Furthermore, while these studies use gaze recordings of native readers, our work focuses 
on non-native readers.

\textbf{NLI in production} NLI was first introduced in Koppel et al. \shortcite{koppel2005} and has 
been drawing considerable attention in NLP, including a recent shared-task challenge with 29 participating teams 
\cite{nli2013report}. NLI has also been driving much of the work on identification of native language 
related features in writing \cite{tsur2007using,jarvis2012approaching,brooke2012,tetreault2012native,swanson2013,swanson2014,malmasi2014,bykh2016}. 
Several studies have also linked usage patterns and grammatical errors in production to linguistic
properties of the writer's native language \cite{nagata13,nagata14,berzak2014,berzak2015}.
Our work departs from NLI in writing and introduces NLI and related feature analysis in 
reading.

\section{Conclusion and Outlook}
\label{sec:conclusion}

We present a novel framework for studying cross-linguistic influence in multilingualism 
by measuring gaze fixations during reading of free-form English text. 
We demonstrate for the first time that this signal can be used to determine 
a reader's native language. The effectiveness of linguistically motivated criteria for fixation
clustering and our subsequent analysis suggest that the ESL reading process
is affected by linguistic factors. Specifically, we show that linguistic similarities 
between native languages are reflected in similarities in ESL reading. We also identify several key features 
that characterize reading in different native languages, and discuss their 
potential connection to structural and lexical properties of the native langauge. The presented results 
demonstrate that eyetracking data can be instrumental for developing
predictive and explanatory models of second language reading. 

While this work is focused
on NLIR from fixations, our general framework can be used to address additional 
aspects of reading, such as analysis of saccades and gaze trajectories. In future work, we 
also plan to explore the role of native and second language writing system characteristics 
in second language reading. 
More broadly, our methodology introduces parallels with production studies in NLP, creating new 
opportunities for integration of data, methodologies and tasks between production and comprehension. 
Furthermore, it holds promise for formulating language learning theory that is supported by 
empirical findings in naturalistic setups across language processing domains.

\section*{Acknowledgements}
We thank Amelia Smith, Emily Weng, Run Chen and Lila Jansen 
for contributions to stimuli preparation and data collection.
We also thank Andrei Barbu, Guy Ben-Yosef, Yen-Ling Kuo, Roger Levy, Jonathan Malmaud, Karthik Narasimhan
and the anonymous reviewers for valuable feedback on this work. This material 
is based upon work supported by the Center for Brains, Minds, and Machines (CBMM), 
funded by NSF STC award CCF-1231216.

\bibliography{acl2017}

\begin{thebibliography}{}
\expandafter\ifx\csname natexlab\endcsname\relax\def\natexlab#1{#1}\fi

\bibitem[{Alonso(2015)}]{alonso2015}
Rosa~Alonso Alonso. 2015.
\newblock {\em Crosslinguistic Influence in Second Language Acquisition\/},
  volume~95.
\newblock Multilingual Matters.

\bibitem[{Barrett et~al.(2016)Barrett, Bingel, Keller, and
  S{\o}gaard}]{barrett2016pos}
Maria Barrett, Joachim Bingel, Frank Keller, and Anders S{\o}gaard. 2016.
\newblock Weakly supervised part-of-speech tagging using eye-tracking data.
\newblock In {\em ACL\/}. volume~2, pages 579--584.

\bibitem[{Barrett and S{\o}gaard(2015{\natexlab{a}})}]{barrett2015pos}
Maria Barrett and Anders S{\o}gaard. 2015{\natexlab{a}}.
\newblock Reading behavior predicts syntactic categories.
\newblock In {\em CoNLL\/}. pages 345--349.

\bibitem[{Barrett and S{\o}gaard(2015{\natexlab{b}})}]{barrett2015parsing}
Maria Barrett and Anders S{\o}gaard. 2015{\natexlab{b}}.
\newblock Using reading behavior to predict grammatical functions.
\newblock In {\em Proceedings of the Sixth Workshop on Cognitive Aspects of
  Computational Language Learning\/}. pages 1--5.

\bibitem[{Berkes and Flynn(2012)}]{berkes2012multilingualism}
{\'E}va Berkes and Suzanne Flynn. 2012.
\newblock Multilingualism: New perspectives on syntactic development.
\newblock {\em The Handbook of Bilingualism and Multilingualism, Second
  Edition\/} pages 137--167.

\bibitem[{Berzak et~al.(2014)Berzak, Reichart, and Katz}]{berzak2014}
Yevgeni Berzak, Roi Reichart, and Boris Katz. 2014.
\newblock Reconstructing native language typology from foreign language usage.
\newblock In {\em Eighteenth Conference on Computational Natural Language
  Learning (CoNLL)\/}.

\bibitem[{Berzak et~al.(2015)Berzak, Reichart, and Katz}]{berzak2015}
Yevgeni Berzak, Roi Reichart, and Boris Katz. 2015.
\newblock Contrastive analysis with predictive power: Typology driven
  estimation of grammatical error distributions in esl.
\newblock In {\em Conference on Computational Natural Language Learning
  (CoNLL)\/}.

\bibitem[{Brooke and Hirst(2012)}]{brooke2012}
Julian Brooke and Graeme Hirst. 2012.
\newblock Measuring interlanguage: Native language identification with
  l1-influence metrics.
\newblock In {\em LREC\/}. pages 779--784.

\bibitem[{Bykh and Meurers(2016)}]{bykh2016}
Serhiy Bykh and Detmar Meurers. 2016.
\newblock Advancing linguistic features and insights by label-informed feature
  grouping: An exploration in the context of native language identification.
\newblock In {\em COLING\/}.

\bibitem[{Byrd et~al.(1995)Byrd, Lu, Nocedal, and Zhu}]{lbfgs1995}
Richard~H Byrd, Peihuang Lu, Jorge Nocedal, and Ciyou Zhu. 1995.
\newblock A limited memory algorithm for bound constrained optimization.
\newblock {\em SIAM Journal on Scientific Computing\/} 16(5):1190--1208.

\bibitem[{Collins and Kayne(2009)}]{sswl}
Chris Collins and Richard Kayne. 2009.
\newblock {\em Syntactic Structures of the world's languages\/}.
\newblock
  \href{http://sswl.railsplayground.net}{http://sswl.railsplayground.net}.

\bibitem[{Cop et~al.(2015)Cop, Drieghe, and Duyck}]{novel15}
Uschi Cop, Denis Drieghe, and Wouter Duyck. 2015.
\newblock Eye movement patterns in natural reading: A comparison of monolingual
  and bilingual reading of a novel.
\newblock {\em PLOS ONE\/} 10(8):1--38.

\bibitem[{Dryer and Haspelmath(2013)}]{wals}
Matthew~S. Dryer and Martin Haspelmath, editors. 2013.
\newblock {\em WALS Online\/}.
\newblock Max Planck Institute for Evolutionary Anthropology, Leipzig.
\newblock \href{http://wals.info/}{http://wals.info/}.

\bibitem[{Dussias(2010)}]{dussias2010uses}
Paola~E Dussias. 2010.
\newblock Uses of eye-tracking data in second language sentence processing
  research.
\newblock {\em Annual Review of Applied Linguistics\/} 30:149--166.

\bibitem[{Hammarstr{\"o}m et~al.(2015)Hammarstr{\"o}m, Forkel, Haspelmath, and
  Bank}]{hammarstrom2015glottolog}
Harald Hammarstr{\"o}m, Robert Forkel, Martin Haspelmath, and Sebastian Bank.
  2015.
\newblock \href{http://glottolog.org}{Glottolog 2.6}.
\newblock {\em Leipzig: Max Planck Institute for Evolutionary Anthropology.\/}
  \href{http://glottolog.org}{http://glottolog.org}.

\bibitem[{Jarvis and Crossley(2012)}]{jarvis2012approaching}
Scott Jarvis and Scott~A Crossley. 2012.
\newblock {\em Approaching Language Transfer Through Text Classification:
  Explorations in the Detection-based Approach\/}, volume~64.
\newblock Multilingual Matters.

\bibitem[{Jarvis and Pavlenko(2008)}]{jarvis2008}
Scott Jarvis and Aneta Pavlenko. 2008.
\newblock {\em Crosslinguistic influence in language and cognition\/}.
\newblock Routledge.

\bibitem[{Klerke et~al.(2016)Klerke, Goldberg, and
  S{\o}gaard}]{klerke2016compression}
Sigrid Klerke, Yoav Goldberg, and Anders S{\o}gaard. 2016.
\newblock Improving sentence compression by learning to predict gaze.
\newblock {\em NAACL-HLT\/} .

\bibitem[{Koppel et~al.(2005)Koppel, Schler, and Zigdon}]{koppel2005}
Moshe Koppel, Jonathan Schler, and Kfir Zigdon. 2005.
\newblock Determining an author's native language by mining a text for errors.
\newblock In {\em Proceedings of the eleventh ACM SIGKDD international
  conference on Knowledge discovery in data mining\/}. ACM, pages 624--628.

\bibitem[{Lewis et~al.(2015)Lewis, Simons, and Fennig}]{ethnologue-2015}
Paul~M. Lewis, Gary~F. Simons, and Charles~D. Fennig, editors. 2015.
\newblock {\em Ethnologue: Languages of the World\/}.
\newblock SIL International, Dallas, Texas.
\newblock \href{http://www.ethnologue.com}{http://www.ethnologue.com}.

\bibitem[{Littel et~al.(2016)Littel, Mortensen, and Levin}]{uriel}
Patrick Littel, David Mortensen, and Lori Levin, editors. 2016.
\newblock {\em URIEL Typological Database\/}.
\newblock Pittsburgh: Carnegie Mellon University.
\newblock
  \href{http://www.cs.cmu.edu/~dmortens/uriel.html}{http://www.cs.cmu.edu/~dmortens/uriel.html}.

\bibitem[{Littell et~al.(2017)Littell, Mortensen, Lin, Kairis, Turner, and
  Levin}]{littell2017lang2vec}
Patrick Littell, David Mortensen, Ke~Lin, Katherine Kairis, Carlisle Turner,
  and Lori Levin. 2017.
\newblock Uriel and lang2vec: Representing languages as typological,
  geographical, and phylogenetic vectors.
\newblock {\em EACL 2017\/} page~8.

\bibitem[{Malmasi and Dras(2014)}]{malmasi2014}
Shervin Malmasi and Mark Dras. 2014.
\newblock Language transfer hypotheses with linear svm weights.
\newblock In {\em EMNLP\/}. pages 1385--1390.

\bibitem[{Marcus et~al.(1993)Marcus, Marcinkiewicz, and Santorini}]{marcus1993}
Mitchell~P Marcus, Mary~Ann Marcinkiewicz, and Beatrice Santorini. 1993.
\newblock Building a large annotated corpus of english: The penn treebank.
\newblock {\em Computational linguistics\/} 19(2):313--330.

\bibitem[{Martohardjono and Flynn(1995)}]{martohardjono1995}
Gita Martohardjono and Suzanne Flynn. 1995.
\newblock Language transfer: what do we really mean.
\newblock In L.~Eubank, L.~Selinker, and M.~Sharwood~Smith, editors, {\em The
  current state of Interlanguage: studies in honor of William E. Rutherford\/},
  John Benjamins: The Netherlands, pages 205--219.

\bibitem[{McDonald et~al.(2013)McDonald, Nivre, Quirmbach-Brundage, Goldberg,
  Das, Ganchev, Hall, Petrov, Zhang, T{\"a}ckstr{\"o}m
  et~al.}]{mcdonald2013universal}
Ryan~T McDonald, Joakim Nivre, Yvonne Quirmbach-Brundage, Yoav Goldberg,
  Dipanjan Das, Kuzman Ganchev, Keith~B Hall, Slav Petrov, Hao Zhang, Oscar
  T{\"a}ckstr{\"o}m, et~al. 2013.
\newblock Universal dependency annotation for multilingual parsing.
\newblock In {\em ACL\/}. pages 92--97.

\bibitem[{Nagata(2014)}]{nagata14}
Ryo Nagata. 2014.
\newblock Language family relationship preserved in non-native english.
\newblock In {\em COLING\/}. pages 1940--1949.

\bibitem[{Nagata and Whittaker(2013)}]{nagata13}
Ryo Nagata and Edward W.~D. Whittaker. 2013.
\newblock Reconstructing an indo-european family tree from non-native english
  texts.
\newblock In {\em ACL\/}. pages 1137--1147.

\bibitem[{Odlin(1989)}]{odlin1989}
Terence Odlin. 1989.
\newblock {\em Language transfer: Cross-linguistic influence in language
  learning\/}.
\newblock Cambridge University Press.

\bibitem[{Petrov et~al.(2012)Petrov, Das, and McDonald}]{petrov2012}
Slav Petrov, Dipanjan Das, and Ryan McDonald. 2012.
\newblock A universal part-of-speech tagset.
\newblock In {\em LREC\/}.

\bibitem[{Piantadosi et~al.(2011)Piantadosi, Tily, and Gibson}]{piantadosi2011}
Steven~T Piantadosi, Harry Tily, and Edward Gibson. 2011.
\newblock Word lengths are optimized for efficient communication.
\newblock {\em Proceedings of the National Academy of Sciences\/}
  108(9):3526--3529.

\bibitem[{Roberts and Siyanova-Chanturia(2013)}]{roberts2013using}
Leah Roberts and Anna Siyanova-Chanturia. 2013.
\newblock Using eye-tracking to investigate topics in l2 acquisition and l2
  processing.
\newblock {\em Studies in Second Language Acquisition\/} 35(02):213--235.

\bibitem[{Santorini(1990)}]{ptbpos}
Beatrice Santorini. 1990.
\newblock Part-of-speech tagging guidelines for the penn treebank project (3rd
  revision).
\newblock {\em Technical Reports (CIS)\/} .

\bibitem[{Swanson and Charniak(2013)}]{swanson2013}
Ben Swanson and Eugene Charniak. 2013.
\newblock Extracting the native language signal for second language
  acquisition.
\newblock In {\em HLT-NAACL\/}. pages 85--94.

\bibitem[{Swanson and Charniak(2014)}]{swanson2014}
Ben Swanson and Eugene Charniak. 2014.
\newblock Data driven language transfer hypotheses.
\newblock {\em EACL\/} page 169.

\bibitem[{Tetreault et~al.(2013)Tetreault, Blanchard, and
  Cahill}]{nli2013report}
Joel Tetreault, Daniel Blanchard, and Aoife Cahill. 2013.
\newblock A report on the first native language identification shared task.
\newblock In {\em Proceedings of the Eighth Workshop on Innovative Use of NLP
  for Building Educational Applications\/}. Citeseer, pages 48--57.

\bibitem[{Tetreault et~al.(2012)Tetreault, Blanchard, Cahill, and
  Chodorow}]{tetreault2012native}
Joel~R Tetreault, Daniel Blanchard, Aoife Cahill, and Martin Chodorow. 2012.
\newblock Native tongues, lost and found: Resources and empirical evaluations
  in native language identification.
\newblock In {\em COLING\/}. pages 2585--2602.

\bibitem[{Tsur and Rappoport(2007)}]{tsur2007using}
Oren Tsur and Ari Rappoport. 2007.
\newblock Using classifier features for studying the effect of native language
  on the choice of written second language words.
\newblock In {\em Proceedings of the Workshop on Cognitive Aspects of
  Computational Language Acquisition\/}. Association for Computational
  Linguistics, pages 9--16.

\bibitem[{Ward~Jr(1963)}]{ward1963}
Joe~H Ward~Jr. 1963.
\newblock Hierarchical grouping to optimize an objective function.
\newblock {\em Journal of the American statistical association\/}
  58(301):236--244.

\bibitem[{Yannakoudakis et~al.(2011)Yannakoudakis, Briscoe, and
  Medlock}]{fce2011}
Helen Yannakoudakis, Ted Briscoe, and Ben Medlock. 2011.
\newblock A new dataset and method for automatically grading esol texts.
\newblock In {\em ACL\/}. pages 180--189.

\end{thebibliography}
\bibliographystyle{acl_natbib}

\appendix

\section{Supplemental Material}
\label{sec:supplemental}

\textbf{Eyetracking Setup} We use a 44.5x30cm screen with 1024x768px resolution 
to present the reading materials, and a desktop mount Eyelink 1000 eyetracker (1000Hz) to record gaze. 
The screen, eyetracker camera and chinrest are horizontally aligned on a table surface. 
The screen center (x=512, y=384) is 79cm away from the center of the forehead bar, and 
13cm below it. The eyetracker camera knob 
is 65cm away from forehead bar. Throughout the experiment participants hold a joystick 
with a button for indicating sentence completion, and two buttons for answering yes/no questions.
We record gaze of the participant's dominant eye.

\textbf{Text Parameters} All the textual material in the experiment is presented using Times font, normal style,
with font size 23. In our setup, this corresponds to 0.36 degrees (11.3px) average lower case letter width, 
and 0.49 degrees (15.7px) average upper case letter width. We chose a non-monospace font, as such fonts
are generally more common in reading. They are also more compact compared to monospace fonts,
allowing to substantially increase the upper limit for sentence length. 

\textbf{Calibration} We use 3H line calibration with point repetition on the central 
horizontal line (y=384), using 16px outer circle, 6px inner circle, fixation points. 
At least three calibrations are performed during the experiment, one at the beginning of each experimental section. 
We also recalibrate upon failure to produce a 300ms fixation 
on any fixation trigger preceding a sentence or a question within 4 seconds after its appearance.
The mean validation error for calibrations across subjects is 0.146 degrees (std 0.038).

\end{document}